\documentclass[letterpaper, 10 pt, conference]{ieeeconf}  % Comment this line out if you need a4paper

\IEEEoverridecommandlockouts                              % This command is only needed if 
\usepackage{amsmath} % assumes amsmath package installed
\usepackage{amssymb}  % assumes amsmath package installed

\newcommand{\vm}[1]{\mathbf{#1}}

\newcommand{\eg}{\textit{e}.\textit{g}.}

\usepackage[pagebackref,breaklinks,colorlinks,citecolor=blue]{hyperref}

\title{\LARGE \bf
Elite-EvGS: Learning Event-based 3D Gaussian Splatting by Distilling Event-to-Video Priors
}

\author{Zixin Zhang$^{1}$, Kanghao Chen$^{1}$ and Lin Wang$^{1,2,*}$% <-this % stops a space
\thanks{*Corresponding author}% <-this % stops a space
\textit{\thanks{$^{1}$ Z. Zhang and K. Chen are with the AI Thrust, The Hong Kong University of Science and Technology (Guangzhou), Guangdong, China.
        {\tt\small zixinzhang@hkust-gz.edu.cn}, 
        {\tt\small kchen879@connect.hkust-gz.edu.cn}
        }%
\thanks{$^{1,2}$Lin Wang is with AI/CMA Thrust, HKUST(GZ) and Dept. of CSE, HKUST, Hong Kong SAR, China, Email:
        {\tt\small linwang@ust.hk}}%}
        }
        }

\usepackage{graphicx}
\usepackage{booktabs}
\usepackage{multirow}
\usepackage{float}

\begin{document}

\maketitle
\thispagestyle{empty}
\pagestyle{empty}

%%%%%%%%%%%%%%%%%%%%%%%%%%%%%%%%%%%%%%%%%%%%%%%%%%%%%%%%%%%%%%%%%%%%%%%%%%%%%%%%
\begin{abstract}
Event cameras are bio-inspired sensors that output asynchronous and sparse event streams, instead of fixed frames. Benefiting from their distinct advantages, such as high dynamic range and high temporal resolution, event cameras have been applied to address 3D reconstruction, important for robotic mapping. 
Recently, neural rendering techniques, such as 3D Gaussian splatting (3DGS), have been shown successful in 3D reconstruction. However, it still remains under-explored how to develop an effective event-based 3DGS pipeline. 
In particular, as 3DGS typically depends on high-quality initialization and dense multiview constraints, a potential problem appears for the 3DGS optimization with events given its inherent sparse property.
To this end, we propose a novel event-based 3DGS framework, named \textbf{Elite-EvGS}.
Our key idea is to distill the prior knowledge from the off-the-shelf event-to-video (E2V) models to effectively reconstruct 3D scenes from events in a coarse-to-fine optimization manner. Specifically, to address the complexity of 3DGS initialization from events, we introduce a novel \textit{warm-up initialization strategy} that optimizes a coarse 3DGS from the frames generated by E2V models and then incorporates events to refine the details. Then, we propose a \textit{progressive event supervision strategy} that
employs the window-slicing operation to 
progressively reduce the number of events used for supervision. This subtly relives the temporal randomness of the event frames, benefiting the optimization of local textural and global structural details.
Experiments on the benchmark datasets demonstrate that Elite-EvGS can reconstruct 3D scenes with better textural and structural details. Meanwhile, our method yields plausible performance on the captured real-world data, including diverse challenging conditions, such as fast motion and low light scenes. For demo and more results, please check our project page: \url{https://vlislab22.github.io/elite-evgs/}.
% The code and dataset will be publicly available.

\end{abstract}

%%%%%%%%%%%%%%%%%%%%%%%%%%%%%%%%%%%%%%%%%%%%%%%%%%%%%%%%%%%%%%%%%%%%%%%%%%%%%%%%
\section{Introduction}

Event cameras, known for their high dynamic range, high temporal resolution, and low latency, have emerged as promising sensors for 3D reconstruction of scenes, which is important for a range of robotic vision tasks, such as visual odometry (VO)~\cite{hadviger2021feature,kueng2016low,zhou2021event}, SLAM~\cite{kim2016real,rebecq2018emvs,zhou2018semi} and navigation~\cite{mahlknecht2022exploring}, manipulation~\cite{funk2024evetac}.
% , and depth estimation~\cite{gehrig2021combining,pan2024srfnet}. 
Unlike conventional frame-based cameras, event cameras capture pixel-wise intensity changes asynchronously, providing an alternative approach to motion analysis~\cite{stoffregen2019event,zhou2021event} and dynamic scene understanding~\cite{kong2024openess,chen2023segment}. These properties make event cameras beneficial for 3D reconstruction in scenarios involving high-speed motion and challenging lighting conditions. 

% Recently, with the advancements in 3D reconstruction techniques such as Neural Radiance Fields (NeRF) and 3D Gaussian Splatting (3DGS), event cameras have been employed for 3D reconstruction—a foundational task for many robotic applications, including navigation~\cite{mahlknecht2022exploring}, manipulation~\cite{funk2024evetac}, and 3D understanding~\cite{chen2022ecsnet}. Notably, existing methods~\cite{eventnerf,qi2023e2nerf} have made progress in reconstructing NeRFs using event cameras. However, due to the slow rendering speed of NeRF, 3DGS-based methods~\cite{xiong2024event3dgs,deguchi2024e2gs} have been proposed to improve efficiency and rendering quality. Despite these advances, these approaches require both RGB frames and event data, which requires systematic alignment between the both sensors. Moreover, they primarily rely on RGB frames, using event data only as supplementary information. This restriction hinders the event camera from fully realizing its potential in extreme scenarios where RGB cameras may completely fail. To the best of our knowledge, no effective 3DGS pipeline exists that relies solely on event data, highlighting a gap in the current research.

With recent advancements in 3D reconstruction techniques such as Neural Radiance Fields (NeRF) and 3D Gaussian Splatting (3DGS), they have been employed for 3D reconstruction with event cameras.
% a foundational task for many robotic applications, including navigation~\cite{mahlknecht2022exploring}, manipulation~\cite{funk2024evetac}, and 3D understanding~\cite{chen2022ecsnet}.
Existing methods~\cite{eventnerf,qi2023e2nerf} have made progress in event-based NeRF reconstruction. However, due to the slow rendering speed of NeRF, 3DGS-based approaches~\cite{xiong2024event3dgs,deguchi2024e2gs} have been introduced to enhance efficiency and rendering quality. 
Despite these advances, these methods struggle with two key challenges: \textbf{1)} they
require both RGB frames and event data, necessitating the systematic alignment of the two sensors; \textbf{2)} they primarily depend on RGB frames, treating event data as an auxiliary modality. This limitation restricts the full potential of event cameras, particularly in extreme scenarios, where RGB cameras may fail. To the best of our knowledge, \textit{there is no effective 3DGS pipeline that relies solely on event data for 3D reconstruction in diverse visual conditions, revealing a significant gap in the current research.}

\begin{figure}[t]
    \centering
    \includegraphics[width=1\linewidth]{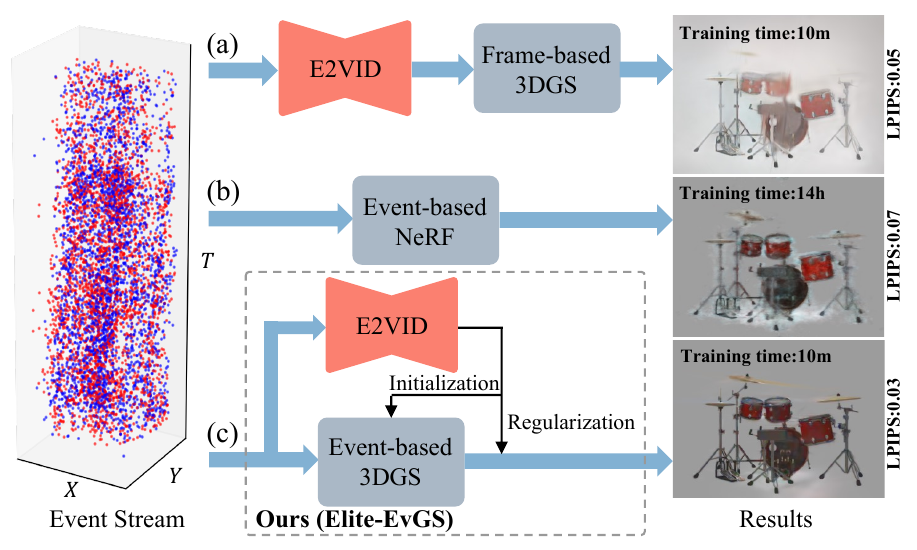}
     \vspace{-20pt}
    \caption{Methodological and qualitative comparison among the E2VID-based, EventNeRF, and our proposed Elite-EvGS. (a) A baseline that employs the E2V models first reconstructs video frames from events and then generates the 3D scene. (b) EventNeRF~\cite{eventnerf}, on the other hand, relies solely on event loss to learn a NeRF representation. (c) We propose an event-based 3DGS approach that distills the geometric priors from the E2VID models to effectively reconstruct 3D scenes from events (with better LPIPS scores and lower training time than EventNeRF).}
    \vspace{-15pt}
    \label{fig:teaser_figure}
\end{figure}

% To fill this gap, we introduce \textbf{Stable-EvGS}, a novel 3D reconstruction framework that utilizes only event streams as input to reconstruct 3DGS. This is a non-trivial problem, given the unique characteristics of event data and intricate of 3DGS. Firstly, since event cameras only record changes in light intensity, the spatial information in event data is very sparse, which makes the 3DGS optimization process under-constrained and unstable. Additionally, 3DGS relies heavily on a high-quality initialization process~\cite{relaxing}, especially in cases where constraints are insufficient. Moreover, due to the randomness in the richness of scene textures and the speed of camera movement, the 
% multi-view event stream exhibits imbalanced temporal distribution. A large time window for stacking events may fail to recover detailed textures due to fast camera movement, while a small time window may struggle to reconstruct the global structure.

% To fill this gap, we introduce \textbf{Stable-EvGS}, a novel 3D reconstruction framework that utilizes only event streams to reconstruct 3DGS. 

This is a non-trivial challenge due to the unique characteristics of event cameras and the complexity of 3DGS optimization. Firstly,  event data is sparse, making the 3DGS optimization under-constrained and unstable. Meanwhile, 3DGS heavily relies on high-quality initialization~\cite{relaxing}, particularly in scenarios with insufficient geometric constraints. 
Moreover, the randomness in scene texture richness and variations in camera movement speed cause the multi-view event stream to have an imbalanced temporal distribution, leading to difficulty in optimization. A large time window for stacking events may fail to capture detailed textures in cases of fast camera movement, while a smaller time window may struggle to recover the global structure accurately.

Previously, event-to-video (E2V) models,\eg~\cite{e2v,ercan2024hypere2vid} have been introduced to bridge event-based and frame-based vision. As a bonus, a naive two-stage optimization baseline is to directly use the pre-trained E2V models to reconstruct multi-view images first and train 3D Gaussians, as illustrated in Fig.~\ref{fig:teaser_figure} (a). Unfortunately, since the E2V models introduce cumulative errors, such a baseline struggles to capture fine details. Despite the limitation, it can reconstruct the scene with a coarse structure, highlighting that E2V models can still provide crucial structural priors. This motivates us to explore combining the best of both worlds -- E2V models and 3DGS-- by distilling the prior knowledge from E2V models to develop an effective event-based 3DGS pipeline that can reconstruct 3D scenes under diverse visual conditions. 

% In this paper, we propose a hybrid pipeline by integrating the E2V model and event-based optimization for a more stable reconstruction, as shown in Fig~\ref{fig:teaser_figure} (c). Specifically, to compensate for the optimization difficulties caused by the sparsity of event data, we leverage pre-trained E2VID models to reconstruct corresponding video frames, using them to train a coarse 3DGS. It provides a stable way to capture global structure, albeit with limited detail. Following this, we jointly optimize both the dense event data and the reconstructed frames to refine the details of the scenes. This hybrid approach fully leverages the prior from the E2V model and raw information of event data, considering global structure while also preserving local details. Considering the imbalanced temporal distribution of event data, we further introduce a progressive window-slicing technique considering event count, facilitating an adaptive event loss.

To this end, we propose a novel event-based 3DGS framework that distills prior knowledge from the E2V model to the event-based optimization for more stable 3D reconstruction, as shown in Fig~\ref{fig:teaser_figure} (c). Specifically, to address the challenges posed by the sparsity of event data as well as the complexity of the optimization of 3D Gaussians, we propose a novel warm-up initialization strategy (Sec.~\ref{sec:e2v}). It utilizes pre-trained E2V models to reconstruct video frames from events, and then uses these frames to train a coarse 3DGS. This enables optimizing a coarse 3DGS -- a foundation for capturing the rough structure of a 3D scene, albeit with limited texture details. Subsequently, we jointly incorporate the event data to refine scene details, where event data serves as the primary supervision and the reconstructed frames serve as a regularization. 
In a nutshell, our warm-up initialization effectively leverages the priors from the E2V model to learn an event-based 3DGS that balances the global structure and preserves the fine details. 
Meanwhile, to address the imbalanced temporal distribution of event data, we introduce a progressive event supervision strategy (Sec.~\ref{sec:event_optimization}) that slices the temporal windows based on the event count instead of time. This makes it possible to progressively reduce the number of events for supervision, benefiting a more effective optimization for the local textural and global structural details.
% We comprehensively evaluate Stable-EvGS on both synthetic and real-world datasets. Extensive experiments demonstrate that our method achieves state-of-the-art performance in 3D reconstruction from only event data, and provides a stable and efficient solution towards diverse extreme conditions.

% Our main contributions are as follows:
% \begin{itemize}
%     \item We present the first effective pipeline for reconstructing 3DGS using only event data, by facilitating a hybrid pipeline by distilling the power from E2V models.
%     \item We propose to utilize the frames reconstructed from E2V models for initialization and the optimize the 3DGS based on event data and regularize with the frames.
%     \item An additional progressive window-slicing technique is proposed to stabilize the optimization process.
% \end{itemize}
    % \item We thoroughly evaluate our method on synthetic and real-world datasets, demonstrating the high-quality and robustness of our approach, which .

We comprehensively evaluate Elite-EvGS on both synthetic and real-world datasets. Extensive experiments demonstrate that our method achieves state-of-the-art performance in 3D reconstruction using only event data, providing a stable and efficient solution for handling diverse and extreme conditions. 
Our main contributions are as follows: (\textbf{I}) We introduce an effective pipeline for reconstructing 3DGS based on pure events, named Elite-EvGS; (\textbf{II}) We propose a novel strategy that distill knowledge from event-to-video models by using frames reconstructed from E2V models for initialization and additional regularization. (\textbf{III}) We introduce a progressive event supervision strategy to further stabilize the optimization process, enhancing reconstruction quality.
% \begin{itemize}
%     \item 
%     \item 
%     \item 
% \end{itemize}

%%%%%%%%%%%%%%%%%%%%%% Related Works %%%%%%%%%%%%%%%%%%%%%%%%%
\section{Related Works}

\noindent \textbf{Event-based NeRF/3DGS:}
% Event cameras have demonstrated significant advantages in various computer vision tasks, including novel view synthesis. 
Works such as EventNeRF~\cite{eventnerf} and Ev-Nerf~\cite{hwang2023ev} leverage event data to synthesize novel views in scenarios involving high-speed motion, which is difficult to capture with conventional cameras. However, these methods assume temporally dense, low-noise event streams—conditions that are often unrealistic in practical applications. To address this, Robust e-NeRF~\cite{low2023robust} incorporates a more realistic event generation model, enabling direct and robust NeRF reconstruction under diverse real-world conditions. Also, DE-NeRF~\cite{hou2024nerf} and E2NeRF~\cite{qi2023e2nerf} extend event-based NeRF approaches to dynamic scenes and severely blurred images, aligning with recent advancements in NeRF research.
Recently, researchers have been exploring 3DGS for reconstructing appearance and geometry from events~\cite{xiong2024event3dgs,deguchi2024e2gs}, 
% leading to significant improvements in reconstruction speed and quality. 
However,  these methods struggle with two key challenges: 1) they require both RGB frames and event data, necessitating the systematic alignment of the two sensors; 2) they primarily
depend on RGB frames, treating event data as an auxiliary modality. This limitation restricts the full potential of event cameras, particularly in extreme scenarios, where RGB cameras may fail.  However, the absence of texture details in event data limits the effectiveness of these approaches. \textit{By contrast, we propose an event-based 3DGS pipeline that distills the prior knowledge from the pre-trained E2V models to effectively reconstruct 3D scenes from events}.

% In this paper, to facilitate high-quality 3DGS reconstruction with clear textures, we incorporate priors from pre-trained E2VID models.

\noindent \textbf{Event-to-Video Reconstruction:}
Recently, deep learning methods~\cite{rebecq2019high,rebecq2019events,scheerlinck2020fast,liu2023sensing,ercan2024hypere2vid} have shown the potential to solve the event-based video reconstruction problem, with the development of advanced network architecture and high-quality data.
% While the previous model-based approaches have been proposed to exploit the relationship between events and intensity frames, recent learning-based methods have shown the potentials to solve the event-based video reconstruction problem 
% Existing methods can be categorized into model-based and learning-based approaches. Model-based methods~\cite{bardow2016simultaneous,munda2018real,brandli2014real,scheerlinck2018continuous} exploit the relationship between events and intensity frames through hand-crafted regularization techniques, however, their results are generally inferior compared to more recent learning-based methods~\cite{rebecq2019high,rebecq2019events,scheerlinck2020fast,liu2023sensing,ercan2024hypere2vid}.
For instance, E2VID~\cite{rebecq2019high,rebecq2019events} employs a U-Net-like network with skip connections and ConvLSTM units to reconstruct videos from long event streams. SPADE-E2VID~\cite{cadena2021spade} further improves temporal coherence by integrating previously reconstructed frames into a SPADE block. HyperE2VID~\cite{ercan2024hypere2vid} introduces hypernetworks that generate per-pixel adaptive filters, guided by a context fusion module. 
% GANs~\cite{wang2019event,yu2020event} and diffusion models~\cite{chen2024lase} have also been applied to address this problem. 
In this paper, we find that E2V models can reconstruct the scene
with a coarse structure, highlighting their potential for providing crucial structural priors. Accordingly, \textit{we aim to distil the prior knowledge from the off-the-shelf E2V models to effectively reconstruct 3D scenes from events.}
% in
% a coarse-to-fine optimization manner. 

% Previous 3D reconstruction methods typically treat E2VID as a two-stage pipeline—first reconstructing videos from events, and then using those videos to reconstruct the 3D scene. However, none of these approaches have explored the possibility of unifying these two tasks into a single pipeline, which could combine the strengths of both event-to-video reconstruction and 3D reconstruction for improved performance.

%%%%%%%%%%%%%%%%%%%%%% Methods %%%%%%%%%%%%%%%%%%%%%%%%%
\section{Methods}

% We present Elite-EvGS, a method for reconstructing 3DGS using only event data. The overall pipeline of our approach is illustrated in Fig.~\ref{fig:pipeline}. Specifically, we first introduce the fundamentals of 3DGS in Sec.~\ref{sec:preliminary}, followed by a detailed explanation of how prior knowledge is distilled from E2V models in Sec.~\ref{sec:e2v}. Moreover, the progressive event-based optimization is described in Sec.~\ref{sec:event_optimization}.

\begin{figure*}[t]
    \centering
    \includegraphics[width=1\linewidth]{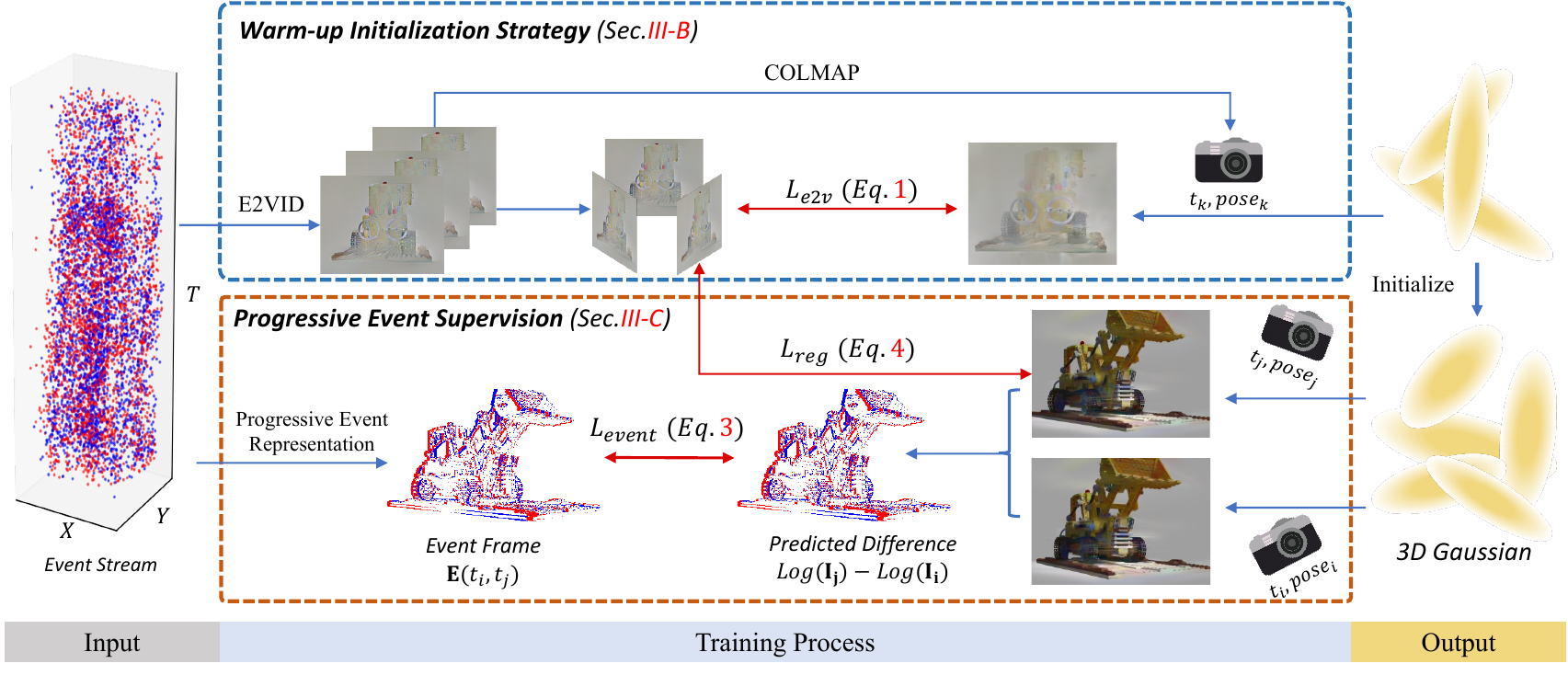}
    \vspace{-22pt}
    \caption{\textbf{Overview of our Elite-EvGS framework}. It takes the event stream as input and outputs a trained set of Gaussians. We first utilize E2VID models to initialize 3D Gaussians (see Sec.~\ref{sec:e2v}). Then, we propose an adaptive event loss to supervise 3D Gaussians with events directly. (see Sec.~\ref{sec:event_optimization}).}
    \vspace{-10pt}
    \label{fig:pipeline}
\end{figure*}

\noindent \textbf{Preliminary.}
\label{sec:preliminary}
% This section introduces the preliminary knowledge about GS.
3D GS~\cite{3dgs} -- an explicit representation of 3D scenes -- employs anisotropic 3D Gaussian functions to represent static 3D scenes. For each Gaussian function, it can be defined as $G(x)=e^{-\frac{1}{2} (x-\mu)^T \Sigma^{-1} (x-\mu)}.\label{eq1}$, 
% \begin{equation}
% \end{equation}
where $\Sigma \in \mathbb{R}^{3 \times 3}$ is an anisotropic covariance matrix,  $\mu \in \mathbb{R}^3$ is its mean vector, also known as the center point of the Gaussian. The covariance matrix $\Sigma$ is parameterized by a scale matrix $S$ and a rotation matrix $R$, $\Sigma= R S S^T R^T.\label{eq2}$
And then the matrix $S$ is stored by its diagonal elements $s=\operatorname{diag}\left(s_x, s_y, s_z\right)$, and the rotation matrix $R$ is constructed from a unit quaternion $r \in \mathbb{R}^4$.
3D Gaussians also include opacity $\sigma\in \mathbb{R}$ and spherical harmonics (SH) coefficients for representing view-dependent colors $c \in \mathbb{R}^3$.

During rendering, 3D Gaussians must be projected into 2D image space~\cite{zwicker2001ewa}. Using the viewing transformation $W$ and the Jacobian of the affine approximation of the projective transformation $J$, the 2D covariance matrix in camera coordinates can be calculated by
$\Sigma^{\prime}=J W \Sigma W^T J^T.\label{eq3}$
Then, the color $C(p)$ of the pixel on the image plane, denoted by $p$, is computed by blending $N$ visible Gaussians sorted by depth~\cite{3dgs}, which is formulated as: $C(p)=\sum_{i \in N} c_i \alpha_i \prod_{j=1}^{i-1}\left(1-\alpha_i\right).\label{eq4}$, where $\alpha_i = \sigma_{i} e^{-\frac{1}{2}(p-\mu_i)^T {\Sigma^{\prime}}^{-1} (p-\mu_i)}$.
% \begin{equation}
% C(p)=\sum_{i \in N} c_i \alpha_i \prod_{j=1}^{i-1}\left(1-\alpha_i\right).\label{eq4}
% \end{equation}
% where $\alpha_i = \sigma_{i} e^{-\frac{1}{2}(p-\mu_i)^T {\Sigma^{\prime}}^{-1} (p-\mu_i)}$.
% % \begin{equation}
%     \label{equ:alpha}
% \end{equation}
Finally, under the supervision of rendering loss, the attributes of each Gaussian are optimized, including position $\mu$, color $c$ represented by spherical harmonics coefficients, opacity $\alpha$, unit quaternion $q$ and scale factor $s$. 
During the optimization process, colmap~\cite{colmap} for initialization andadaptive density control\cite{3dgs} are used for a stable and fine reconstruction.
% However, although 3DGS surpass NeRF on training speed and rendering quality, the discrete and explicit representation makes the pipeline vulnerable to the superparameter and difficult to optimize. For event data, the sparsity further increases the difficulty of reconstruction.
\subsection{Overview}
An overview of the our Elite-EvGS is illustrated in Fig.~\ref{fig:pipeline}.
It distills the prior knowledge from the pretrained E2V models to effectively reconstruct 3D scenes from events in a coarse-to-fine optimization manner. Specifically, to address the complexity of 3DGS initialization from sparse events, we introduce the \textit{warm-up initialization strategy} (Sec.~\ref{sec:e2v}) that optimizes a coarse 3DGS from the frames generated by E2V models and then incorporates events to refine the details. Then, we propose the \textit{progressive event supervision strategy} (Sec.~\ref{sec:event_optimization}) to alleviate the temporal randomness of the event frames, benefiting for the optimization of event-based 3DGS. We now describe the technical details.

% that
% % hat 
% employs the window-slicing operation to 
% progressively reduce the number of events used for supervision. 
% This subtly relives the temporal randomness of the event frames, benefiting for the optimization of local texture and global structure details.

\subsection{Warm-up Initialization Strategy}
\label{sec:e2v}
% Although 3DGS\cite{3dgs} surpass NeRF\cite{nerf} in both training speed and rendering quality, the discrete and explicit representation makes the pipeline vulnerable to the super parameters and difficult to optimize. 
The sparsity of event data complicates the reconstruction process. Since event cameras only record pixels that exhibit brightness changes, the sparse information they provide is often insufficient for effective supervision. This results in an ill-posed optimization problem, unless there is significant camera movement in a highly textured environment.

\begin{figure}[t!]
    \centering
    \includegraphics[width=1\linewidth]{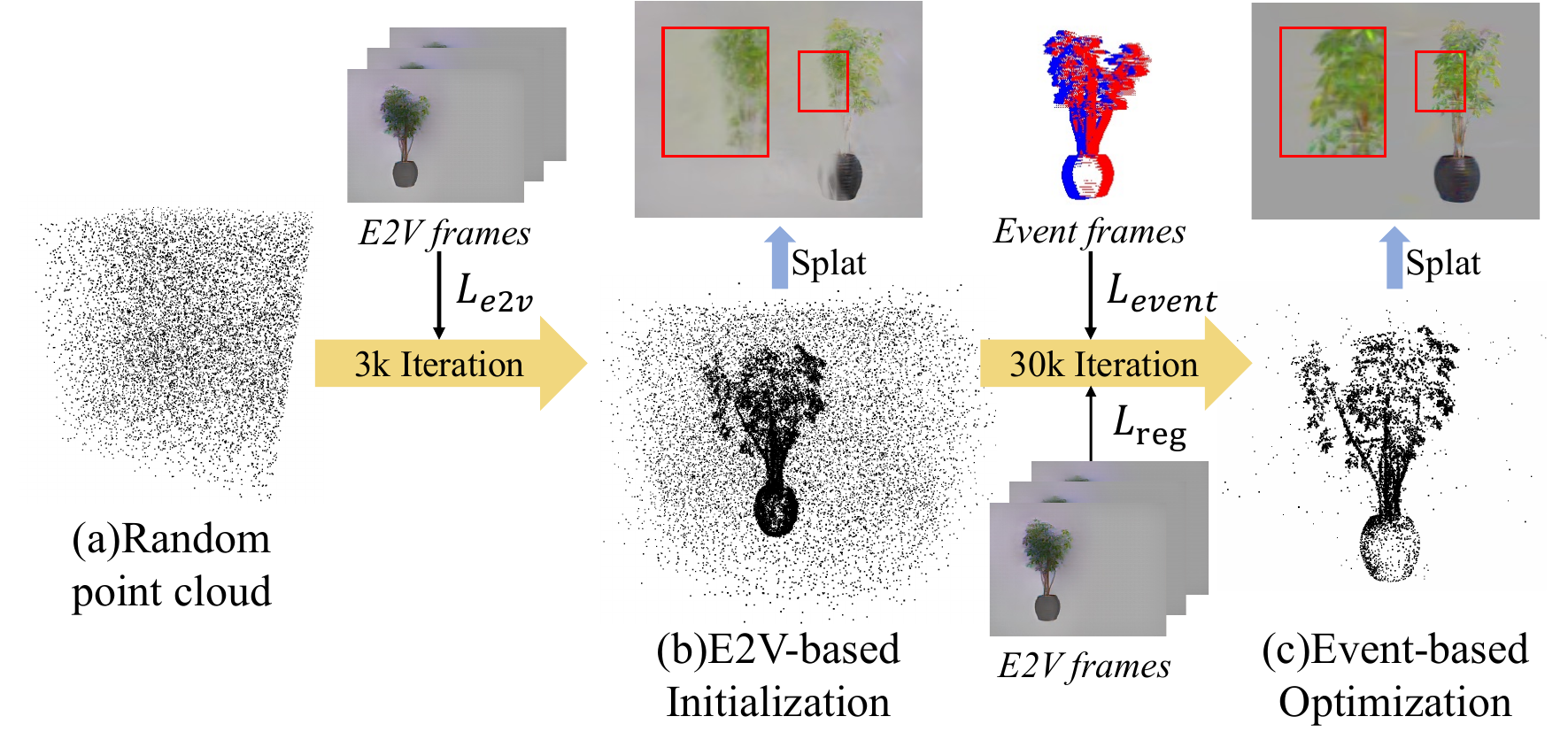}
       \vspace{-22pt}
    \caption{Illustration and visualization of our warm-up initialization strategy.}
    \vspace{-10pt}
    \label{fig:e2v-init}
\end{figure}

To address these issues, we propose the warm-up initialization strategy to distill prior knowledge from E2V models. We utilize these models to generate images from the event stream. Although the resulted images are noisy and may not accurately represent colors, they retain substantial structural information. We extract and refine this prior knowledge through an initialization process and a regularization term. The optimization process of the 3DGS includes three stages:

% \subsubsection{Initiliazation}

\noindent \textbf{Random Point Could.} As is illustrated in Fig.~\ref{fig:e2v-init} (a), we start with a randomly initialized 3D Gaussian point cloud. 

\noindent \textbf{E2V-based Initialization.} 
Then, we use an image-based rendering loss to optimize the 3D Gaussians for 3k itreation:
\begin{equation}
    \mathcal{L}_{e2v} = ||\vm{I}_{e2v} - \vm{I}_{render}||_1,
\end{equation}
where $\vm{I}_{e2v}$ represents the reconstructed frames from the E2VID model and $\vm{I}_{render}$ represents the rendered image from the 3D Gaussian. 
By performing our warm-up initialization strategy, we can obtain a coarse 3D Gaussian point cloud, as shown in Fig.~\ref{fig:e2v-init}(b).
Although the resulting point cloud exhibits considerable noise, it successfully recovers the approximate geometric structure of the object, providing an initial state for further optimization.
% However, using this loss function throughout the entire optimization process can lead to overfitting on noise and errors present in the E2V images, resulting in suboptimal outcomes. To mitigate this, we perform a warm-up for initialization and apply early stopping after a fixed number of steps. 

\begin{figure*}[t!]
    \centering
    \includegraphics[width=.85\linewidth]{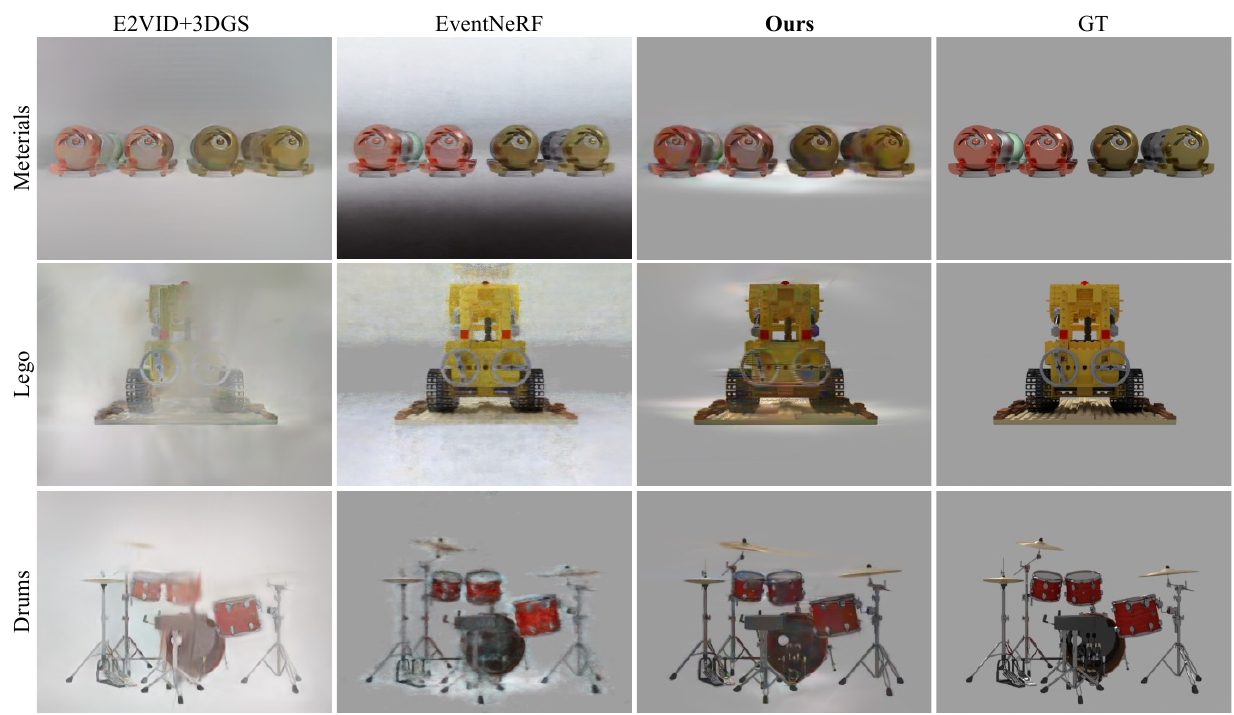}
    \vspace{-10pt}
    \caption{\textbf{Qualitative comparison of different methods on EventNeRF Dataset}~\cite{eventnerf}. Our method outperforms the baseline, including EventNeRF and E2VID~\cite{e2v}+3DGS~\cite{3dgs} with better textural and geometric details.}
    \vspace{-5pt}
    \label{fig:eventnerf-comp}
\end{figure*}

\begin{table*}[t]
    \caption{\textbf{Quantitative comparison on EventNeRF\cite{eventnerf} dataset.}}
    \vspace{-10pt}
    \centering
    \begin{tabular*}{0.8\linewidth}{c|ccc|ccc|ccc}
    \toprule
    \multirow{2}{*}{Scene} &
    \multicolumn{3}{c|}{EventNeRF\cite{eventnerf}} & 
    \multicolumn{3}{c|}{E2VID+3DGS} & 
    \multicolumn{3}{c}{\textbf{Ours}} \\     
    \cmidrule(lr){2-4} \cmidrule(lr){5-7} \cmidrule(lr){8-10}
   & PSNR $\uparrow$ & SSIM $\uparrow$ & LPIPS $\downarrow$ & PSNR $\uparrow$ & SSIM $\uparrow$ & LPIPS $\downarrow$ & PSNR $\uparrow$ & SSIM $\uparrow$ & LPIPS $\downarrow$ \\ 
   \midrule
    Drums                  & 27.43           & 0.91            & 0.07               
    & 23.57           & 0.94            & 0.05
    & \textbf{29.37}           & \textbf{0.96}            &\textbf{0.04}                    \\
    Lego                   & 25.84           & 0.89            & 0.13               
    & 22.90           & 0.91            & 0.08
    & \textbf{26.97}           & \textbf{0.94}            &\textbf{0.05}               \\
    Chair                  & 30.62           & 0.94            & 0.05               
    & 28.74           & 0.98            & 0.04
    & \textbf{31.86}           & \textbf{0.98}            & \textbf{0.03}               \\
    Ficus                  & 31.94           & 0.94            & 0.05               
    & 31.25           & 0.98           & 0.02
    & \textbf{40.80}           & \textbf{0.99}            & \textbf{0.01}               \\
    Mic                    & 31.78           & 0.96            & 0.03               
    & 36.20           & 0.99            & 0.01
    & \textbf{36.31}           & \textbf{0.99}            & \textbf{0.01}               \\
    Hotdog                 & 30.26           & 0.94            & 0.04               
    & 24.33           & 0.96            & 0.04
    & \textbf{30.60}           & \textbf{0.97}            & \textbf{0.03}               \\
    Materials              & 24.10           & 0.94            & 0.07               
    & 26.51           & 0.95            & 0.05
    & \textbf{27.13}           & \textbf{0.95}            & \textbf{0.04}               \\ 
    \hline
    Average                & 28.85           & 0.93            & 0.06               
    & 27.64                & 0.96                & 0.04
    & \textbf{31.86}           & \textbf{0.97}            & \textbf{0.03}               \\
    \bottomrule
    \end{tabular*}
    \label{tab:EventNeRF}
\end{table*}

\noindent \textbf{Event-based Optimization.} After initialization, we use the original event data with a regularization term to supervise the 3D Gaussians, for an ultimate fine 3D Gaussians as shown in Fig.\ref{fig:e2v-init}(c). Detailed elaboration of the optimization process at this stage can be found in Sec. \ref{sec:event_optimization}.
.

\subsection{Progressive Event Supervision Strategy}
\label{sec:event_optimization}
\noindent \textbf{Progressive Event Representation.} Considering the CUDA-based rasterization of the 3DGS pipeline renders the entire image, we process the event stream into a frame-based representation to better suit the pipeline of 3DGS. Specifically, given two arbitrary timestamps $t_1$ and $t_2$, we integrate the events within this time interval and map them onto pixels to form an event frame, which is formulated as:  
\begin{equation}
\label{eq:event}
    E_{x,y}(t_1, t_2) = \sum_{(\tau,p) : t_1 < \tau \leq t_2} p\Delta,
\end{equation}
where $p$ is the polarity of the event, and $\Delta$ is the systematic threshold of event camera.
Due to variations in scene textures and camera movement speeds, the temporal density of events exhibits randomness. Using a fixed time interval for segmentation can result in an uneven distribution of events~\cite{wang2019event, rebecq2019high,gallego2020event,zheng2023deep}. Therefore, we adopt a strategy that segments event frames based on event count, rather than fixed time intervals. 
The time interval $\Delta t$ between $t_1$ and $t_2$ is determined by a specified number $k$, such that the total number of triggered events within this interval is $k$.
% \begin{equation}
%     t_2 - t_1 = \Delta t, where \Delta t = \sum_{(\tau,p) : t_1 < \tau \leq t_1 + \Delta t} |p| = k
% \end{equation}
When supervised with a large number of events, the model tends to learn the overall structure of the scene, whereas with fewer events, it focuses more on the finer details. To achieve a coarse-to-fine training process, we progressively reduce the number $k$.

\noindent \textbf{Event Supervision Loss.}  During the optimization process, we render two image $\vm{I}_{t_1}$, $\vm{I}_{t_2}$ at timestamps $t_1$, $t_2$. We then convert $\vm{I}_{t_1}$ and $\vm{I}_{t_2}$ into log space and compute the difference between the two. This difference map is compared with the ground truth event frame to calculate a loss for supervision.
\begin{equation}
    \mathcal{L}_{event} = ||(\log(\vm{I}_{t_1}) - \log(\vm{I}_{t_2})), \vm{E}(t_1, t_2)||_2,
\end{equation}
where $\vm{E}$ represents the event frame with elements of  ${E_{x,y}}$ in position of $(x,y)$.

\begin{table*}[t!]
    \caption{\textbf{Quantitative comparison on Real-World dataset.}}
     \vspace{-10pt}
    \centering
    \begin{tabular*}{0.8\linewidth}{c|ccc|ccc|ccc}
    \toprule
    \multirow{2}{*}{Scene} &
    \multicolumn{3}{c|}{EventNeRF\cite{eventnerf}} & 
    \multicolumn{3}{c|}{E2VID+3DGS} & 
    \multicolumn{3}{c}{\textbf{Ours}} \\     
    \cmidrule(lr){2-4} \cmidrule(lr){5-7} \cmidrule(lr){8-10}
   & PSNR $\uparrow$ & SSIM $\uparrow$ & LPIPS $\downarrow$ & PSNR $\uparrow$ & SSIM $\uparrow$ & LPIPS $\downarrow$ & PSNR $\uparrow$ & SSIM $\uparrow$ & LPIPS $\downarrow$ \\ 
   \midrule
    Normal                  & 23.55           & 0.72            & 0.47    
    & 27.82           & 0.79            & 0.42
    & \textbf{28.20}           & \textbf{0.80}            &\textbf{0.41}   \\
    Blur                  & 17.60           & 0.67            & 0.50 
    & 24.02           & 0.79            & 0.36
    & \textbf{24.50}           & \textbf{0.80}            &\textbf{0.35}   \\
    Low light                  & 21.33           & 0.65            & 0.47
    & 23.07           & 0.68            & 0.40               
    & \textbf{23.91}           & \textbf{0.71}            &\textbf{0.40}   \\
    \hline
    Average                & 20.83                & 0.68           & 0.48
    & 24.97           & 0.75            & 0.39               
    & \textbf{25.54}           & \textbf{0.77}            & \textbf{0.39}               \\
    \bottomrule
    \end{tabular*}
    \label{tab:real}
\end{table*}

\begin{figure*}[t!]
    \centering
    \includegraphics[width=.85\linewidth]{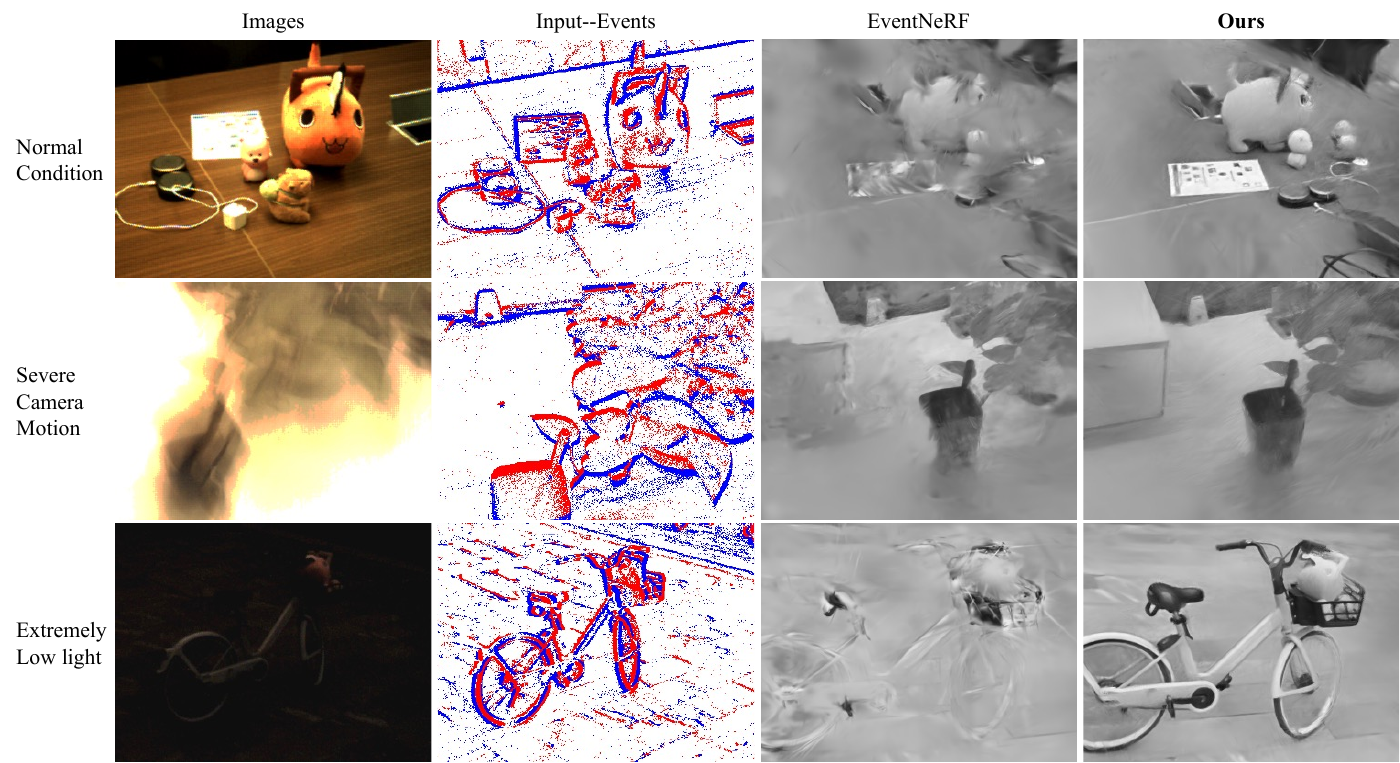}
     \vspace{-10pt}
    \caption{\textbf{Qualitative comparison on our captured real-world dataset}. The input is event data. We compared with EventNeRF regarding the ability to synthesize novel views. Our method yields better real-world novel view synthesis capacity than EventNeRF.}
    \label{fig:real-comp}
\end{figure*}

\noindent \textbf{E2V Regularization Loss.}
Moreover, we introduce a regularization term that leverages structural information from E2V prior while mitigating the overfitting to image noise, imposing a weaker constraint on the optimization process. This approach helps event-based optimization achieve a more stable outcome.
Specifically, we base the regularization term on the Structural Similarity Index (SSIM), which evaluates the perceptual quality of an image by assessing brightness, contrast, and structure:
\begin{align}
\label{eq:reg}
\mathcal{L}_{reg} = 1 - SSIM(\vm{I}_{e2v}, \vm{I}_{render}).
\end{align}
This loss further extracts useful information from the E2VID prior and supports event-based optimization, facilitating a more stable optimization process.

\noindent \textbf{Total Loss.} The final optimization loss is given by:
\begin{equation}
\mathcal{L} = \lambda_{event} \mathcal{L}_{event} + \lambda_{reg} \mathcal{L}_{reg},
\end{equation}
where $\lambda_{event}$ is set to 0.02 and $\lambda_{reg}$ is set to 0.002.
%%%%%%%%%%%%%%%%%%%%%% Experiments %%%%%%%%%%%%%%%%%%%%%%%%%

\section{Experiments}
\subsection{Dataset and Metrics, and Implementation Details}
\subsubsection{Dataset} 
We evaluate our method using both synthetic and real-world data. 
\textbf{Synthetic Blender Dataset:} For synthetic scenes, we adopt the dataset proposed in EventNeRF~\cite{eventnerf}, which generates 7 sequences with 360$^{\circ}$ camera rotations around each 3D object in Blender, simulating event streams from 1000 views.
% \subsubsection{Simulated Real-World Dataset}Given the difficulty in obtaining ground truth images using an actual event camera, we have employed the v2e model to generate event streams from videos shot in real-world scenes, allowing us to create paired data for quantitative evaluation of our method. Specificaly, we ...
\textbf{Real-world Dataset:} We captured 3 sequences using the DAVIS346 Event Camera on the indoor and outdoor scenes, encompassing conditions with normal light, low light, and rapid camera motion. 
% For normal light condition, we use APS frames as ground truth to quantitatively evaluate our method's ability of novel view synthesis. For 

\subsubsection{Metrics} We use three metrics to evaluate our method: PSNR, SSIM~\cite{wang2004image}, and LPIPS~\cite{zhang2018unreasonable} to evaluate the similarity between synthesized novel view and the reference. Following EventNeRF~\cite{eventnerf}, we apply a linear transformation in the logarithmic space for all our baseline results.
\subsubsection{Implementation Details}
We implement our method based on the official 3DGS\cite{3dgs}, retaining the default hyper-parameters. During training, we first start from a random point cloud with 10,000 points and apply warm-up initialization as in Sec.\ref{sec:e2v}. For the E2V model, we use the pre-trained E2VID\cite{e2v} model for inference. For adaptive event frame supervision, the event count $k$ was progressively reduced from 150,000 to 30,000 during the optimization process. All experiments were done on a RTX 3090 GPU.

\subsection{Results on Synthetic Datasets}
We conducted both quantitative and qualitative comparisons on EventNeRF Dataset. We compared three methods, the EventNeRF~\cite{eventnerf}, the E2VID~\cite{e2v}+3DGS ~\cite{3dgs} and our Elite-EvGS model. For EventNeRF, we use its original code and trained for 500k iteration under their default setting.

Quantitative and qualitative comparisons are shown in Tab.\ref{tab:EventNeRF} and Fig. \ref{fig:eventnerf-comp}, respectively. 
Our method achieves state-of-the-art (SOTA) performance on all the metrics for novel view synthesis. Specifically, our method shows a 3.01 dB(10\%) improvement in PSNR, 0.04 (4\%) improvement in SSIM and 0.03 (50\%) improvement in LPIPS.
As shown in Fig. \ref{fig:eventnerf-comp}, compared to the EventNeRF method and our baseline, our approach reconstructs objects that are significantly clearer and have less noise.
Moreover, compared to EventNeRF, our method is 80 times faster in training speed thanks to the computational efficency of 3DGS.
% and significantly improves reconstruction quality
% , achieving state-of-the-art (SOTA) performance on the PSNR, SSIM, and LPIPS metrics for novel viewpoint images.

\subsection{Results on Our Captured Real-World Datasets}
We have also verified our framework under more realistic and extreme conditions, as illustrated in the Fig~\ref{fig:real-comp}. We evaluated on three scenarios: normal scenes, scenes with intense camera movement, and extreme low-light conditions. For the latter two methods, RGB cameras fail entirely. 
However, our framework capitalizes on the unique advantages of event cameras, achieving consistent 3D reconstruction capabilities across all scenarios. 
% The qualitative comparisons are depicted in the Fig.\ref{fig:real-comp}. 
The reconstruction quality of EventNeRF baseline deteriorates significantly in challenging real-world scenarios, while our method achieves stable 3D reconstruction across various scenes, successfully reconstruct both main object and the background.
 
For quantitative comparison, given that RGB cameras are ineffective in extreme conditions, the ground truth images from new viewpoints are unavailable. So we utilize images generated by E2VID at the unseen viewpoint as reference to assess our framework’s capability to synthesize new viewpoints. Qualitative Results further confirm the robustness of our method under these challenging conditions, far surpassing the EventNeRF baseline.

\begin{table}[t!]
\centering
\caption{Ablation results for the E2VID prior.}
\vspace{-10pt}
\begin{tabular}{c|ccc}
\toprule
\textbf{Method} & \textbf{PSNR$\uparrow$} & \textbf{SSIM$\uparrow$} & \textbf{LPIPS$\downarrow$} \\ \hline
w/o E2VID prior & 24.14 & 0.91 & 0.09 \\ 
3k iter E2VID init & 26.57 & 0.93 & 0.06  \\ 
3k iter E2VID init + reg & 26.97 & 0.94 & 0.05\\ 
\bottomrule
\end{tabular}
\label{tab:ablation-e2v}
\end{table}

\begin{figure}[t!]
    \centering
    % \vspace{-10pt}
    \includegraphics[width=1\linewidth]{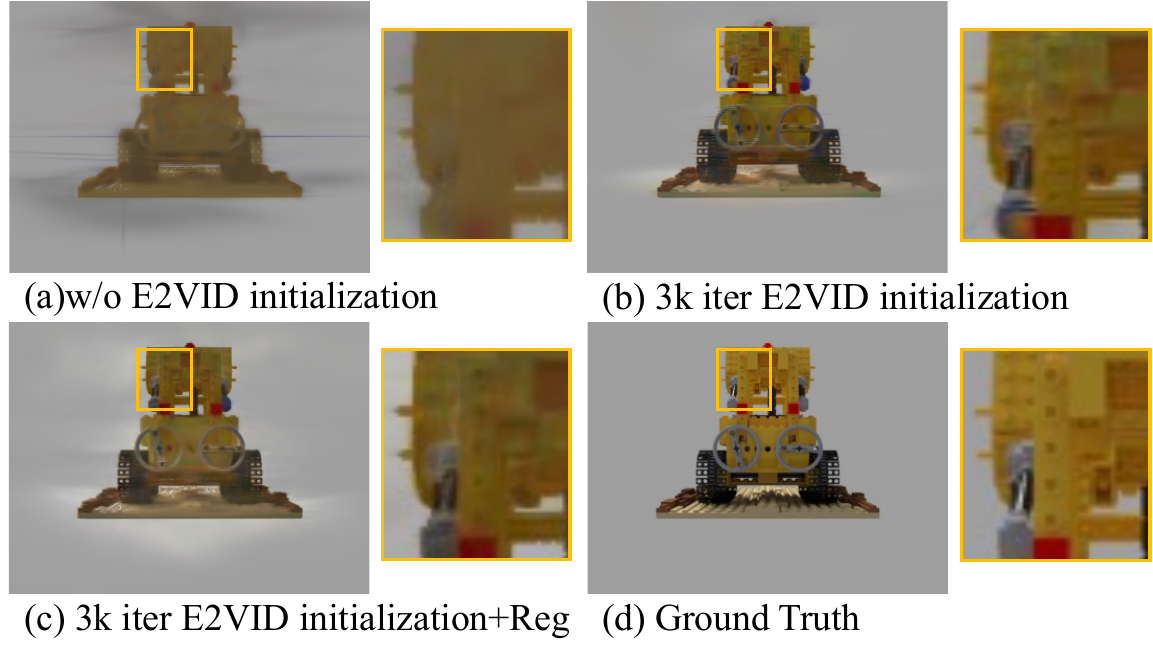}
    \vspace{-20pt}
    \caption{Qualitative results for the ablation on E2VID~\cite{e2v} prior.}
    \vspace{-5pt}
    \label{fig:ablation-e2v}
\end{figure}

\begin{table}[t!]
\centering
\caption{Ablation results on progressive event supervision.}
\vspace{-10pt}
\begin{tabular}{c|ccc}
\toprule
\textbf{Method} & \textbf{PSNR$\uparrow$} & \textbf{SSIM$\uparrow$} & \textbf{LPIPS$\downarrow$} \\ \hline
E2VID + 3DGS& 23.15 & 0.68 & 0.40 \\ 
+ event frame supervision & 23.46 & 0.69 & 0.39 \\ 
+ progressive slicing & 23.91 & 0.71 & 0.39 \\ 
\bottomrule
\end{tabular}
\label{tab:ablation-bike}
\vspace{-5pt}
\end{table}

\begin{figure}[t!]
    \centering
    \includegraphics[width=1\linewidth]{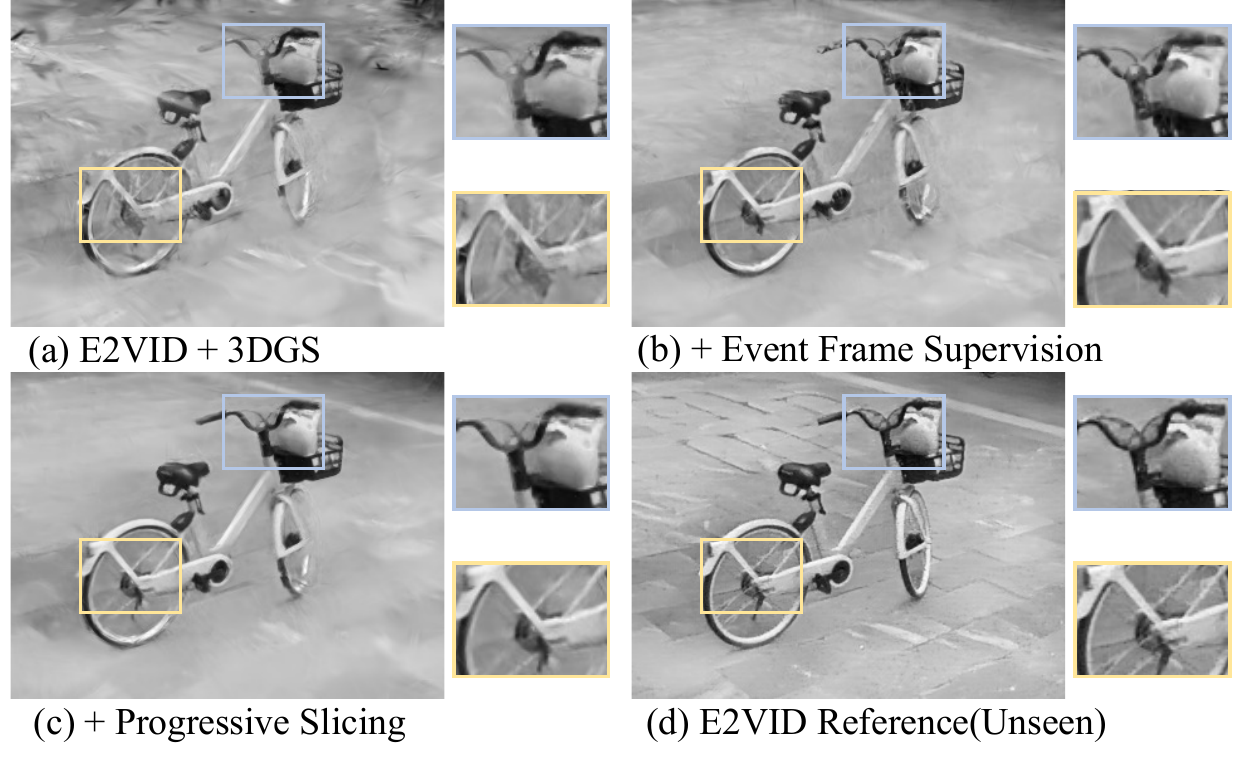}
    \vspace{-20pt}
    \caption{Ablation results on adaptive event frame supervision. With the proposed module, our method more effectively recovers both the structure and details of a scene, including the handlebar (blue box) and wheel (yellow box) components of a bicycle, thus more closely aligning with the unseen E2V reference from a novel view.
}
    \label{fig:ablation-bike}
    \vspace{-5pt}
\end{figure}

\subsection{Ablation Study}
\subsubsection{How does the E2VID prior contribute to the reconstruction process?}
To evaluate the impact of E2VID on the reconstruction process, we performed an ablation study, with the results presented in Fig. \ref{fig:ablation-e2v} and Tab. \ref{tab:ablation-e2v}. We designed three experimental conditions: (a) random initialization without E2VID, (b) 3k iterations with E2VID initialization, and (c) 3k iterations of initialization with an additional regularization term during event-based optimization. Comparing conditions (a) and (b) reveals that the sparsity of event data often leads to local optima in the ill-posed optimization process, resulting in excessively smooth images. However, initializing with E2VID allows the optimization process to converge more quickly to the optimal solution, offering an effective starting point for further optimization. Additionally, the comparison between (b) and (c) shows that the regularization term further enhances geometric accuracy.

\subsubsection{Does Adaptive Event Frame Supervision works?}
To validate the effectiveness of our Adaptive Event Frame Supervision, we conducted an ablation study comparing three scenarios: (a) supervision without events, relying solely on images from E2VID; (b) supervision with a fixed segmentation of event frames after E2V initialization; and (c) supervision employing adaptive and progressive segmentation of event frames after E2V initialization. As is shown in Fig.~\ref{fig:ablation-bike} and Tab.~\ref{tab:ablation-bike}, the experimental results demonstrated the efficacy of event frame supervision and the superiority of our adaptive progressive segmentation approach.

\section{CONCLUSIONS}
% In this paper, we present Elite-EvGS, a novel framework for reconstructing 3DGS from only event data. Using E2V priors for warm-up initialization and employing a progressive slicing technique to facilitate adaptive event loss, Elite-EvGS demonstrates robustness and efficiency in handling the inherent challenges posed by event data. Extensive experiments confirm its state-of-the-art performance across a range of synthetic and real-world scenarios, showcasing its potential for high-quality, event-based 3D reconstruction.

In this paper, we introduce Elite-EvGS, a novel framework for reconstructing 3DGS using only event data. By leveraging E2V priors for warm-up initialization and employing a progressive slicing technique to facilitate adaptive event loss, Elite-EvGS demonstrates robustness and efficiency in addressing the inherent challenges of event data. Extensive experiments validate its state-of-the-art performance across various synthetic and real-world scenarios, highlighting its potential for high-quality, event-driven 3D reconstruction.


%%%%%%%%%%%%%%%%%%%%%%%%%%%%%%%%%%%%%%%%%%%%%%%%%%%%%%%%%%%%%%%%%%%%%%%%%%%%%%%%

%%%%%%%%%%%%%%%%%%%%%%%%%%%%%%%%%%%%%%%%%%%%%%%%%%%%%%%%%%%%%%%%%%%%%%%%%%%%%%%%



% \section*{ACKNOWLEDGMENT}

% The preferred spelling of the word ÒacknowledgmentÓ in America is without an ÒeÓ after the ÒgÓ. Avoid the stilted expression, ÒOne of us (R. B. G.) thanks . . .Ó  Instead, try ÒR. B. G. thanksÓ. Put sponsor acknowledgments in the unnumbered footnote on the first page.

%%%%%%%%%%%%%%%%%%%%%%%%%%%%%%%%%%%%%%%%%%%%%%%%%%%%%%%%%%%%%%%%%%%%%%%%%%%%%%%%

% References are important to the reader; therefore, each citation must be complete and correct. If at all possible, references should be commonly available publications.
\clearpage

\bibliographystyle{plain}
\bibliography{references}

\end{document}